\documentclass[11pt]{article}

\usepackage[preprint]{acl}

\usepackage{times}
\usepackage{latexsym}

\usepackage[T1]{fontenc}

\usepackage[utf8]{inputenc}

\usepackage{microtype}

\usepackage{inconsolata}

\usepackage{graphicx}
\usepackage{amsmath,amssymb,amsfonts,mathtools,amsthm}

\usepackage{graphicx}
\usepackage{subcaption}
\usepackage{booktabs}

\usepackage{microtype}
\usepackage{nicefrac}
\usepackage{url}
\usepackage[table]{xcolor}

\usepackage{wrapfig}
\usepackage{floatflt}
\usepackage{float}
\usepackage{enumitem}
\usepackage{etoc}
\usepackage{pifont}
\usepackage[utf8]{inputenc}
\usepackage[T1]{fontenc}

\usepackage[capitalize,noabbrev]{cleveref}

\usepackage{algorithm}
\usepackage{algpseudocode}
\algrenewcommand\alglinenumber[1]{}

\theoremstyle{plain}

\theoremstyle{definition}

\theoremstyle{remark}

\newcommand{\bl}[1]{\textcolor{black}{#1}}

\usepackage[T1]{fontenc}
\usepackage{multirow}
\usepackage{color, colortbl}
\definecolor{LightCyan}{rgb}{0.93,1,1}
\definecolor{LLightCyan}{rgb}{0.97,1,1}
\definecolor{LightRed}{rgb}{1,0.95,0.95}
\definecolor{LLightRed}{rgb}{1,0.97,0.97}
\definecolor{Gray}{rgb}{0.8,0.8,0.8}
\definecolor{LightGray}{rgb}{0.92,0.92,0.92}
\definecolor{LLightGray}{rgb}{0.95,0.95,0.95}
\definecolor{mygreen}{rgb}{0.4,0.7,0.306}
\definecolor{myblue}{rgb}{0.294,0.447,0.796}
\definecolor{deepgreen}{RGB}{0,100,0}
\usepackage{hhline}

\title{Bayesian Data Reweighting Improves Multimodal Retrieval for Knowledge-Based Visual Question Answering}

\author{
  \textbf{Jingchen Sun\textsuperscript{1}},
  \textbf{Shaobo Han\textsuperscript{2}},
  \textbf{Ruiyi Zhang\textsuperscript{3}},
  \textbf{Naresh Kumar Devulapally\textsuperscript{1}},
  \textbf{Ming Liu\textsuperscript{4}},
\\
  \textbf{Yitao Long\textsuperscript{5}},
  \textbf{Vishnu Suresh Lokhande\textsuperscript{1}},
  \textbf{Changyou Chen\textsuperscript{1}}
\\
\\
  \textsuperscript{1}University at Buffalo,
  \textsuperscript{2}NEC Laboratories America,
  \textsuperscript{3}Adobe Research,
\\
  \textsuperscript{4}Iowa State University,
  \textsuperscript{5}New York University
}

\begin{document}
\maketitle
\begin{abstract}
Multimodal retrievers are essential for knowledge-based visual question answering, where they retrieve external evidence for image–question pairs. However, existing contrastive training methods typically treat all unmatched query–document pairs as equally informative negatives, which is problematic because many unmatched documents may still be semantically relevant or partially useful. We propose \textbf{Bayesian Data Reweighting}, a probabilistic framework that models query–document importance as latent variables and adaptively infers posterior weights to downweight likely false negatives. With closed-form posterior updates under conjugate priors and stochastic EM optimization, our method consistently improves retrieval accuracy across three retrievers and seven knowledge-based VQA benchmarks.

\end{abstract}

\section{Introduction}
Knowledge-based Visual Question Answering (KB-VQA)~\cite{okvqa, schwenk2022okvqa} extends the traditional VQA task by requiring models to incorporate external knowledge sources, such as structured knowledge graphs~\cite{conceptnet}, unstructured textual corpora~\cite{wikidata}, or large-scale encyclopedic documents~\cite{encyclopedic} to answer questions. These questions often involve commonsense reasoning~\cite{commonsensevqa}, fine-grained factual knowledge~\cite{infoseek}, or entity disambiguation~\cite{jian2024llmra}, which is often absent from raw visual or linguistic input. As such, KB-VQA serves as a key benchmark for evaluating a model's ability to integrate perception with world knowledge~\cite{caf2024wikillava, yan2024echosight}, and has significant implications for downstream applications in education, healthcare, and open-domain dialog systems.

\begin{figure}[htbp]
  \centering
\includegraphics[width=0.475\textwidth]{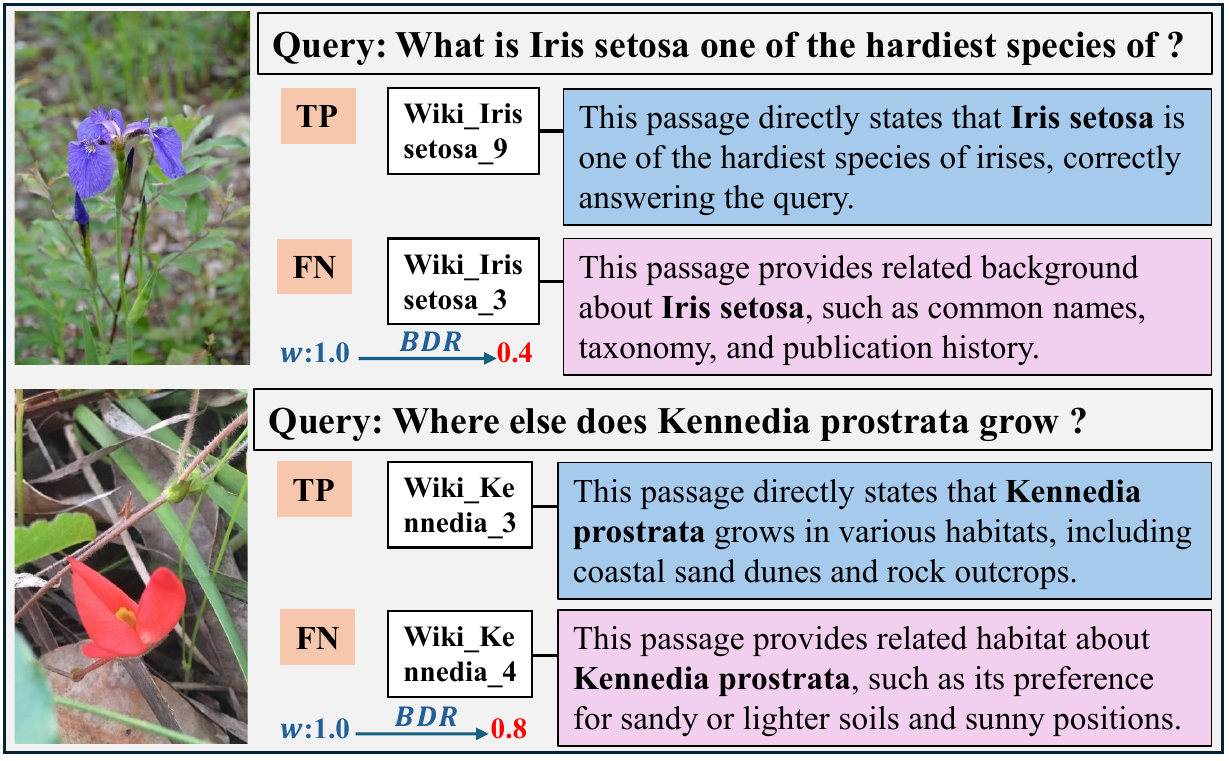}
\caption{\textbf{Examples of false negative reweighting.}
We select representative FN samples from EVQA to illustrate the false-negative issue in KB-VQA, where unpaired documents may still describe related entities or partially relevant evidence. Unlike standard InfoNCE, which uniformly treats these samples as negatives and pushes them away from the query, our method BDR adaptively infers weights for FN passages, suppressing noisy contrastive signals and making the learning process more robust.}
  \label{fig:motivatio}
  \vspace{-16pt}
\end{figure}
Recent advances in knowledge-based visual question answering (KB-VQA) have largely focused on improving the modeling of multimodal retrievers, including late-interaction architectures~\cite{flmr, preflmr}, unified embedding models~\cite{vlm2vec, uniir, mm-embed}, and retrieval-augmented frameworks with stronger generators~\cite{rag-vqa, hu2023reveal}. Despite these architectural progress, most existing retrievers~\cite{flmr, preflmr, recurrence, vlm2vec} are still trained with the standard InfoNCE objective~\cite{infonce}, which treats all non-positive documents in a batch as equally informative negatives.

However, this assumption is particularly problematic in KB-VQA. Unlike conventional image-text retrieval, KB-VQA relies on open-domain external knowledge, where documents are often weakly annotated, semantically overlapping, and only partially aligned with the image-question pair. As illustrated in Figure~\ref{fig:motivatio}, an unpaired document is not necessarily irrelevant: it may describe a closely related entity, provide partially useful evidence, or even support the correct answer but remain unlabeled. Such samples become \textbf{false negatives} when they are incorrectly treated as negatives during contrastive training. Pushing these semantically relevant documents away from the query introduces noisy gradients and can degrade retrieval performance, especially in knowledge-intensive settings where entity ambiguity and incomplete evidence annotations are common.

Prior studies have attempted to address false negatives through negative reweighting or debiased learning methods~\cite{sampling, facenet, decoupled, hard-negative, hard-metric, chuang2020debiased}. However, these methods remain limited for knowledge-intensive retrieval. Sampling or reweighting-based approaches often rely on heuristic scores or predefined rules, which may still assign large weights to semantically relevant but unlabeled documents and amplify noisy supervision. Debiased contrastive learning methods usually apply global correction terms, making it difficult to capture the query-dependent nature of false negatives in KB-VQA. Moreover, existing methods rarely model the uncertainty of whether an unmatched document is truly irrelevant or incorrectly labeled as negative. As a result, they may still push potentially useful evidence away from the query, leading to suboptimal retriever training under incomplete supervision.

To address this limitation, we propose a novel \textbf{Bayesian Data Reweighting (BDR)} framework for noise-aware multimodal retrieval. Instead of assuming all unmatched query--document pairs contribute equally, BDR introduces latent importance weights to model the reliability and usefulness of each pair during training. We formulate contrastive learning as a Bayesian reweighting problem, where sample-wise weights are inferred from posterior distributions under selected priors. This allows the model to adaptively downweight likely false negatives, without explicitly identifying their labels. To make inference tractable and efficient, we derive closed-form posterior updates via auxiliary variable augmentation and optimize the retriever with a stochastic approximation expectation-maximization algorithm. In this way, BDR provides a principled, instance-adaptive training objective that mitigates noisy supervision and improves the robustness of KB-VQA retrieval.

Extensive experiments show that BDR consistently improves multimodal retrieval across seven knowledge-based VQA benchmarks and three different LLM backbones. Compared with standard InfoNCE and existing reweighting baselines, BDR achieves stronger retrieval accuracy and further improves downstream answer generation on InfoSeek and EVQA datasets when paired with frozen generators. Detailed analyses show that BDR learns better-separated query--document embeddings, improves Recall@K across retrieval budgets, and effectively assigns lower weights to annotated false negatives, validating its ability to mitigate noisy supervision. Moreover, BDR incurs negligible additional training cost, making it a practical and scalable objective for robust KB-VQA retrieval.

Our main contributions are summarized as follows: \textbf{First}, we identify the false-negative issue in KB-VQA retrieval, where semantically relevant but unannotated documents are incorrectly treated as negatives under standard contrastive learning. \textbf{Second,} we propose {Bayesian Data Reweighting (BDR)}, a probabilistic framework that models query--document importance as latent variables and adaptively infers posterior weights to suppress noisy false negatives. \textbf{Third}, we conduct extensive experiments across seven KB-VQA benchmarks and three multimodal LLM backbones, demonstrating consistent retrieval improvements, stronger downstream VQA performance, meaningful false-negative-aware weighting, and negligible additional training overhead.

\section{Related Work}
\paragraph{Knowledge-based Visual Question Answering.} 
KB-VQA extends traditional VQA by adopting a retrieval-augmented generation (RAG) paradigm~\cite{okvqa, schwenk2022okvqa}. 
Recent advances in multimodal RAG have demonstrated strong effectiveness for KB-VQA by integrating external textual knowledge into large multimodal language models~\cite{caf2024wikillava, yan2024echosight, long2025reuse}.
Despite these advances, KB-VQA performance remains highly sensitive to retrieval quality~\cite{rag-vqa, revive}. Existing pipelines often retrieve redundant or weakly relevant passages~\cite{hao2024self}, or fail to capture fine-grained entities and attributes grounded in the visual scene~\cite{jian2024llmra}. These limitations motivate us to develop a more robust and effective retriever.

\paragraph{Negative Reweighting in Contrastive Learning.}
False negatives are a common challenge in contrastive learning, where semantically relevant samples are incorrectly treated as negatives. Existing methods typically mitigate this issue by correcting or reducing the contribution of suspicious negatives. For example, debiased contrastive learning~\cite{chuang2020debiased} estimates the true-negative contribution at the distribution level to reduce the bias introduced by false negatives. Other methods identify potential false negatives using sample-level similarity signals and then remove, cancel, or downweight them during training~\cite{boosting-fn,incremental-fn}. Beyond false-negative mitigation, several studies further combine false-negative debiasing with hard-negative reweighting. These methods aim to suppress harmful false negatives while emphasizing informative hard negatives, often by using similarity scores, hardness-aware sampling distributions, margin-based objectives \cite{facenet,hard-negative,hard-mix,hard-metric,approximate-negative}. 

However, most existing methods still rely on heuristic similarity thresholds, predefined sampling rules, or global correction terms. As a result, they may be sensitive to hyperparameter choices and lack a principled mechanism to model the uncertainty of whether an unmatched sample is truly irrelevant or a false negative. This limitation is particularly important in KB-VQA retrieval, where external documents are weakly annotated, semantically dense, and often only partially aligned with the image--question pair. Motivated by this, we propose Bayesian Data Reweighting, which treats sample importance as latent variables and infers posterior weights in an instance-adaptive manner. 

\section{Bayesian Data Reweighting}

Most prior retrievers are trained with the InfoNCE objective~\cite{infonce}. Given a training set
$\mathcal{X}=\{(\mathbf{v}_i,\mathbf{d}_i)\}_{i=1}^{M}$, each pair $(\mathbf{v}_i,\mathbf{d}_i)$ is treated as a positive pair, while $(\mathbf{v}_i,\mathbf{d}_k)$ with $k\neq i$ are treated as negatives. Let $\mathbf{q}_i$ and $\mathbf{z}_k$ denote the query and document embeddings. We define the exponential cosine similarity as
$s_{ik}=\exp\left(\mathrm{sim}(\mathbf{q}_i,\mathbf{z}_k)/\tau\right),$
where $\tau>0$ is the temperature. For a mini-batch $\mathcal{B}$, the standard InfoNCE loss is
\begin{align}
\nonumber
\mathcal{L}(\mathcal{B};\boldsymbol{\theta})
=
-\frac{1}{B}
\sum_{i=1}^{B}
\log
\frac{s_{i^{+}}}
{s_{i^{+}}+\sum_{k=1}^{K}s_{ik^{-}}}.
\end{align}

In KB-VQA retrieval, unmatched documents are not always true negatives, since they may still contain related or partially useful evidence. Inspired by classical importance sampling~\cite{important_sample} and reweighted probabilistic models~\cite{robust}, we introduce a probabilistic variable
$w_{ik}\ge 0$
to model the importance of each negative pair $(\mathbf{v}_i,\mathbf{d}_{ik^-})$.
This leads to the following weighted contrastive objective:
\begin{align}
\nonumber
\mathcal{L}_{\mathrm{BDR}}(\mathcal{B};\boldsymbol{\theta})
=
-\frac{1}{B}
\sum_{i=1}^{B}
\log
\frac{s_{i^{+}}}
{s_{i^{+}}+\sum_{k=1}^{K}w_{ik}s_{ik^{-}}},
\end{align}
where $w_{ik}$ controls the contribution of the negative pair $(\mathbf{v}_i,\mathbf{d}_{ik^-})$. When $w_{ik}=1$ for all negatives, this objective reduces to the standard InfoNCE loss.

\paragraph{Bayesian Modeling of Local Weights}
We treat the negative weights
$
\mathbf{w}
=
\{w_{ik}\}
$
as latent random variables and infer them jointly with the retriever parameters $\boldsymbol{\theta}$ during training, without assuming access to a clean validation set~\cite{ren2018learning} or relying on gradient-based importance estimates~\cite{important_sample}. We interpret the weighted contrastive objective as the following sample-wise pseudo-likelihood:
\[
p_i(\mathbf{w}_i, \boldsymbol{\theta})
\propto
\frac{
s_{i^+}
}{
s_{i^+}
+
\sum_{k=1}^{K}
w_{ik}s_{ik^-}
}.
\]

The joint posterior distribution is
\[
p(\mathbf{w},\boldsymbol{\theta}\mid\mathcal{X})
\propto
\prod_{i=1}^{|\mathcal X|}
p_i(\mathbf{w}_i,\boldsymbol{\theta})
\prod_{i,k}
p(w_{ik})
p(\boldsymbol{\theta}).
\]

Our goal is thus to infer the posterior distribution $p(\mathbf{w},\boldsymbol{\theta}\mid\mathcal{X})$ in a principled Bayesian manner. However, direct Bayesian inference is intractable. The normalization term
$
\lambda_i
=
s_{i^+}
+
\sum_{k=1}^{K}
w_{ik}s_{ik^-}
$
couples all negative weights through a summation.
As a result, the conditional posterior distributions do not admit closed-form solutions under common prior choices.
This breaks the standard conditional conjugacy and prevents direct Gibbs sampling.

\subsection{Laplace-Augmented Factorization}

To address the non-conjugacy caused by the normalization term, we introduce an auxiliary variable $u_i$ using a Laplace-transform-based augmentation technique~\cite{data-augmentation}. Denote the normalization term by
$\lambda_i=s_{i^{+}}+\sum_{k=1}^{K}w_{ik}s_{ik^{-}}$.
Using the identity $\lambda_i^{-1}=\int_{0}^{\infty}\exp(-\lambda_i u_i)\,\mathrm{d}u_i$ for $\lambda_i>0$, the sample-wise pseudo-likelihood admits the following marginally equivalent augmented form:
\begin{align}
\frac{s_{i^{+}}}{\lambda_i}
=
\int_{0}^{\infty}
s_{i^{+}} e^{-\lambda_i u_i}\,\mathrm{d}u_i .
\nonumber
\end{align}
Conditioned on $u_i$, substituting $\lambda_i$ yields the following factorized augmented pseudo-likelihood:
\begin{align}
\widetilde p_i(u_i,\mathbf{w}_i,\boldsymbol{\theta})
\propto
s_{i^+} e^{-u_i s_{i^+}}
\prod_{k=1}^{K} e^{-u_i w_{ik}s_{ik^-}}.
\nonumber
\end{align}
Thus, the original normalized pseudo-likelihood is converted into a conditionally factorized exponential-family form, enabling closed-form conditional posterior updates under conjugate priors.

\paragraph{Posterior Inference via Gibbs Sampling.}
After Laplace augmentation, the latent variables admit closed-form conditional posteriors and can be efficiently sampled by alternating
\[
u_i
\sim
p(u_i\mid\mathbf{w}_i,\boldsymbol{\theta},\mathcal X),
\quad
w_{ik}
\sim
p(w_{ik}\mid u_i,\boldsymbol{\theta},\mathcal X).
\]
Here, $u_i$ acts as an auxiliary normalization variable that decouples the negative weights, making each $w_{ik}$ conditionally independent given $u_i$.

\subsection{Prior Selection and Posterior Inference}

We place a Gamma prior on the auxiliary variable, $u_i\sim \operatorname{Gamma}(a_u,b_u)$, since $u_i>0$ and is conjugate to the exponential factor $e^{-u_i\lambda_i}$ in the augmented pseudo-likelihood. Its conditional posterior is
\[
u_i\mid-
\sim
\operatorname{Gamma}
\left(
a_u,\;
b_u+s_{i^+}+\sum_{k=1}^{K}w_{ik}s_{ik^-}
\right).
\]
For the negative pair weights $w_{ik}$, we consider two prior choices.

\paragraph{Gamma Prior.}
For smooth continuous reweighting, we use a Gamma prior $w_{ik}\sim \operatorname{Gamma}(a_w,b_w)$, where $a_w$ and $b_w$ are the shape and rate parameters. Combining it with the augmented pseudo-likelihood gives
\[
p(w_{ik}\mid-)
\propto
w_{ik}^{a_w-1}
\exp\!\left[-(b_w+u_i s_{ik^-})w_{ik}\right].
\]
and hence
\[
w_{ik}\mid-
\sim
\operatorname{Gamma}
\left(
a_w,\;
b_w+u_i s_{ik^-}
\right).
\]
The expectation of this posterior distribution is
\[
\mathbb{E}[w_{ik}\mid-]
=
\frac{a_w}{b_w+u_i s_{ik^-}}.
\]
By inspecting this posterior expectation, we can observe that: when a negative sample has high similarity to the query, $s_{ik^-}$ increases the posterior rate and decreases the expected weight. Thus, the Gamma prior enables continuous false-negative-aware reweighting.

\paragraph{Bernoulli Prior.}
As an alternative, we use a Bernoulli prior $w_{ik}\sim \operatorname{Bernoulli}(p_w)$ with $w_{ik}\in\{0,1\}$ for binary negative gating, where $w_{ik}=1$ keeps the negative and $w_{ik}=0$ removes it. The unnormalized posterior probabilities are
\[
\begin{aligned}
p(w_{ik}=1\mid-) &\propto p_w e^{-u_i s_{ik^-}},\\
p(w_{ik}=0\mid-) &\propto 1-p_w.
\end{aligned}
\]
After normalization,
\[
w_{ik}\mid-
\sim
\operatorname{Bernoulli}
\left(
\frac{
p_w e^{-u_i s_{ik^-}}
}{
1-p_w+p_w e^{-u_i s_{ik^-}}
}
\right).
\]
The posterior expectation is
\[
\mathbb{E}[w_{ik}\mid-]
=
\frac{
p_w e^{-u_i s_{ik^-}}
}{
1-p_w+p_w e^{-u_i s_{ik^-}}
}.
\]
By inspecting this expectation, we can see that the Bernoulli posterior performs more aggressive false-negative filtering: highly similar negatives have smaller retaining probabilities because $e^{-u_i s_{ik^-}}$ decreases as $s_{ik^-}$ grows. Compared with the Gamma prior, the Bernoulli prior provides hard keep-or-drop gating for likely false negatives.

\subsection{Optimization}
We optimize the proposed \textbf{BDR} framework with stochastic approximation EM (SAEM)~\cite{delyon1999convergence}, shown in Algorithm~\ref{alg:bdr_saem}. At each iteration, BDR samples a mini-batch, computes positive and negative similarities with the current retriever, and draws the auxiliary variable $u_i$ and latent weights $w_{ik}$ from their closed-form conditional posteriors. The sampled weights are then used to form a weighted contrastive loss, which updates the stochastic surrogate objective. Finally, the retriever parameters are optimized by stochastic gradient descent on this surrogate. By alternating posterior inference and parameter updates, BDR adaptively reweights noisy negatives during training.
\algrenewcommand\alglinenumber[1]{}
\begin{algorithm}[H]
\caption{\textbf{BDR} Optimization via SAEM}
\label{alg:bdr_saem}
\begin{algorithmic}[1]
\State \textbf{Input:} training set $\mathcal{X}$, parameters $\boldsymbol{\theta}_0$, step sizes $\{\rho_t\}$, learning rates $\{\eta_t\}$
\State Initialize $Q_0(\boldsymbol{\theta})\gets 0$
\For{$t=0,1,\ldots$}
    \State Sample mini-batch $\mathcal{B}_t=\{(\mathbf{v}_i,\mathbf{d}_i)\}_{i=1}^{B}$
    \State Compute $s_{i^+}$ and $\{s_{ik^-}\}_{k=1}^{K}$ using $\boldsymbol{\theta}_t$
    \Statex \textbf{E-step:}
    \For{each $i\in\mathcal{B}_t$}
        \State Sample $u_i \sim p(u_i\mid \mathbf{w}_i,\boldsymbol{\theta}_t,\mathcal{X})$
        \State Sample $w_{ik}\sim p(w_{ik}\mid u_i,\boldsymbol{\theta}_t,\mathcal{X})$, for $n=1,\ldots,N$
    \EndFor
    \State $\widehat{Q}_{t+1}(\boldsymbol{\theta})\gets \mathcal{L}_{\mathrm{BDR}}(\mathcal{B}_t;\boldsymbol{\theta})$
    \Statex \textbf{M-step:}
    \State $\boldsymbol{\theta}_{t+1}\gets \boldsymbol{\theta}_t-\eta_t\nabla_{\boldsymbol{\theta}}Q_{t+1}(\boldsymbol{\theta})|_{\boldsymbol{\theta}=\boldsymbol{\theta}_t}$
\EndFor
\end{algorithmic}
\end{algorithm}

\begin{table*}[htb]
\centering
\fontsize{8}{8}\selectfont
\setlength{\tabcolsep}{5pt}
\renewcommand{\arraystretch}{1.15}
\caption{\textbf{Retrieval performance comparison on individual dataset.} BDR consistently outperforms uniform sampling and existing reweighting baselines across both Recall@K and Pseudo Recall@K, with the Gamma prior achieving the best overall performance.}
\begin{tabular}{lcccc|cccc}
\toprule
\multirow{2}{*}{Method} 
& \multicolumn{4}{c|}{OKVQA} 
& \multicolumn{4}{c}{EVQA} \\
\cmidrule(lr){2-5} \cmidrule(lr){6-9}
& R@1 & R@5 & PR@1 & PR@5
& R@1 & R@5 & PR@1 & PR@5 \\
\midrule
Uniform Negative Sampling 
& 23.0 & 45.8 & 39.8 & 61.7
& 28.0 & 57.0 & 35.5 & 63.6 \\

Random Negative Reweighting  
& 22.8 & 46.8 & 37.4 & 59.5
& 27.2 & 55.2 & 34.1 & 62.4 \\

Similarity-based Reweighting 
& 24.6 & 47.6 & 39.0 & 60.1
& 27.6 & 58.1 & 34.6 & 64.8 \\

Margin-based Reweighting Loss 
& 25.4 & 49.3 & 40.7 & 61.4
& 28.8 & 59.8 & 35.5 & 65.7 \\

Decoupled Positive Reweighting 
& 25.6 & 49.1 & 40.5 & 60.9
& 29.9 & 60.3 & 36.9 & 67.0 \\

Debiased Negative Reweighting 
& 25.9 & 50.5 & 42.0 & 62.6
& 30.6 & 60.9 & 37.4 & 67.4 \\

Hardness Negative Reweighting 
& 26.5 & 49.8 & 42.6 & 62.8
& 31.7 & 61.1 & 37.6 & 66.9 \\
\hline
Bayesian Reweighting (Bernoulli Prior) 
& $\textbf{28.9}{\scriptstyle \pm 0.7}$ 
& $\textbf{53.8}{\scriptstyle \pm 0.7}$ 
& $\textbf{44.4}{\scriptstyle \pm 1.0}$ 
& $\textbf{63.5}{\scriptstyle \pm 0.4}$ 
& $\textbf{33.5}{\scriptstyle \pm 1.3}$ 
& $\textbf{63.9}{\scriptstyle \pm 1.3}$ 
& $\textbf{40.6}{\scriptstyle \pm 1.3}$ 
& $\textbf{69.7}{\scriptstyle \pm 1.0}$ \\

Bayesian Reweighting (Gamma Prior) 
& $\textbf{30.1}{\scriptstyle \pm 0.4}$ 
& $\textbf{54.8}{\scriptstyle \pm 0.8}$ 
& $\textbf{45.9}{\scriptstyle \pm 0.9}$ 
& $\textbf{66.0}{\scriptstyle \pm 0.6}$ 
& $\textbf{35.1}{\scriptstyle \pm 0.9}$ 
& $\textbf{65.6}{\scriptstyle \pm 1.6}$ 
& $\textbf{42.2}{\scriptstyle \pm 0.9}$ 
& $\textbf{71.5}{\scriptstyle \pm 1.4}$ \\
\bottomrule
\end{tabular}

\label{tab:reweighting_r1_r5_pr1_pr5}
\end{table*}

\begin{table*}[htb]
\centering
\fontsize{8}{8}\selectfont
\setlength{\tabcolsep}{3.5pt}
\renewcommand{\arraystretch}{1.15}
\caption{\textbf{Retrieval performance on M2KR benchmarks.} BDR consistently improves retrieval performance across seven knowledge-intensive VQA datasets, with the Gamma-prior variant achieving the best average Recall.}
\resizebox{\textwidth}{!}{
\begin{tabular}{lcc|cc|cc|cc|cc|cc|cc|c}
\toprule
\multirow{2}{*}{Method}
& \multicolumn{2}{c|}{EVQA}
& \multicolumn{2}{c|}{OKVQA}
& \multicolumn{2}{c|}{KVQA}
& \multicolumn{2}{c|}{InfoSeek}
& \multicolumn{2}{c|}{WIT}
& \multicolumn{2}{c|}{OVEN}
& \multicolumn{2}{c|}{LLaVA}
& \multirow{2}{*}{Avg.} \\
\cmidrule(lr){2-3}
\cmidrule(lr){4-5}
\cmidrule(lr){6-7}
\cmidrule(lr){8-9}
\cmidrule(lr){10-11}
\cmidrule(lr){12-13}
\cmidrule(lr){14-15}
& R@1 & R@5
& R@1 & R@5
& R@1 & R@5
& R@1 & R@5
& R@1 & R@5
& R@1 & R@5
& R@1 & R@5
&  \\
\midrule

Uniform Negative Sampling
& 27.7 & 55.8
& 24.9 & 47.0
& 26.1 & 52.0
& 21.1 & 43.3
& 23.0 & 43.3
& 26.2 & 52.6
& \textbf{77.2} & 92.7
& 43.8 \\

Debiased Negative Reweighting
& 26.9 & 56.7
& 23.1 & 45.4
& 27.1 & 53.6
& 27.8 & 47.7
& 35.6 & 58.5
& 38.7 & 64.2
& 65.3 & 87.2
& 47.0 \\

Hardness Negative Reweighting
& 30.2 & 59.2
& 19.6 & 39.1
& 25.7 & 51.9
& 36.4 & 59.6
& 33.9 & 56.5
& 42.0 & 66.8
& 58.6 & 84.6
& 47.4 \\

\hline

Bayesian Reweighting (Bernoulli Prior)
& 34.0 & 65.1
& 23.5 & 44.0
& 31.9 & 59.1
& \textbf{40.2} & \textbf{66.1}
& 41.7 & 65.1
& 47.4 & 71.5
& 76.0 & \textbf{93.5}
& 54.2 \\

Bayesian Reweighting (Gamma Prior)
& \textbf{37.8} & \textbf{68.8}
& \textbf{26.0} & \textbf{50.0}
& \textbf{35.9} & \textbf{61.6}
& 39.3 & 62.8
& \textbf{42.9} & \textbf{67.0}
& \textbf{48.3} & \textbf{72.7}
& 76.6 & 93.1
& \textbf{55.9} \\

\bottomrule
\end{tabular}
}
\label{tab:multiple_benchmarks}
\end{table*}

\begin{table*}[htb]
\centering
\caption{\textbf{Retrieval performance with different LLM backbones on M2KR.} For OVEN and KVQA, we only report R@5, and for LLaVA, we only report R@1, to ensure comparability with previous baselines: PreFLMR \cite{preflmr} and ReT \cite{recurrence}. BDR consistently improves Qwen2-VL-2B, Phi-3.5-V-3.8B, and Qwen2-VL-7B across knowledge-intensive VQA benchmarks.}
\setlength{\tabcolsep}{4.5pt}
\fontsize{8}{8}\selectfont
\setlength\arrayrulewidth{0.6pt}
\renewcommand{\arraystretch}{1.1}
\begin{tabular}{@{}lcccccccccc@{}}
\toprule
                         & \multicolumn{2}{c}{\textbf{EVQA}}      & \multicolumn{2}{c}{\textbf{OKVQA}}      & \multicolumn{2}{c}{\textbf{InfoSeek}}       & \textbf{OVEN} & \textbf{LLaVA} & \textbf{KVQA} & \textbf{Average}  \\
\hline
\textbf{Retriever}        & \textbf{R@5}  & \textbf{PR@5} & \textbf{R@5}   & \textbf{PR@5} & \textbf{R@5}      & \textbf{PR@5} & \textbf{R@5} & \textbf{R@1}   & \textbf{R@5}  & -    \\
\hline
UniIR (Feature Fusion) \cite{uniir} & 17.0 & 35.5 & 9.3 & 58.1 & 24.5 & 46.0 & 63.6 & 47.8 & 13.5 & 35.0 \\
CLIP (Unimodal) \cite{clip}          & 17.2 & 29.8 & 4.3   & 37.9 & 25.7     & 44.2 & 69.1 & 44.8  & 56.1 & 36.6 \\
CLIP (Feature Fusion) \cite{clip}   & 61.9 & 71.9 & 27.9  & 67.5 & 37.3     & 56.3 & 63.0 & 72.0  & 41.1 & 55.4 \\
PreFLMR \cite{preflmr}       & 62.0 & 72.0 & 30.2 & 67.4 & 39.2 & 57.7 & 64.3 & 72.6 & 41.9 & 56.4 \\
ReT \cite{recurrence}                      & 48.6 & 60.2 & 19.0  & 63.8 & 52.0     & 62.5 & 84.0 & 79.2  & 60.6 & 58.9 \\
\hline
Qwen2-VL-2B        & 50.4 & 63.9 & 24.8  & 58.7 & 58.5 & 53.7 & 75.6 & 84.2 & 51.0 & 57.9 \\
Qwen2-VL-2B w/ BDR     & \textbf{51.2} & \textbf{64.2} & \textbf{26.6}  & \textbf{59.7} & \textbf{60.5} & \textbf{56.7} & \textbf{78.3} & \textbf{88.7} & \textbf{55.6} & \textbf{60.2} \\
\color{blue}{$\Delta$}                 
& \bl{+0.8} & \bl{+0.3} & \bl{+1.8} & \bl{+1.0} 
& \bl{+2.0} & \bl{+3.0} & \bl{+2.7} & \bl{+4.5} 
& \bl{+4.6} & \bl{+2.3}
\\ \hline

Phi-3.5-V-3.8B        & 45.1 & 60.3 & 35.1  & 65.3 & 40.8 & 44.3 & 71.5 & 91.4  & 52.6 & 56.3 \\
Phi-3.5-V-3.8B w/ BDR     & \textbf{49.2} & \textbf{62.2} & \textbf{42.3}  & \textbf{69.1} & \textbf{43.8} & \textbf{47.5} & \textbf{74.7} & \textbf{91.6} & \textbf{57.7} & \textbf{59.8} \\
\color{blue}{$\Delta$}                  
& \bl{+4.1} & \bl{+1.9} & \bl{+7.2} & \bl{+3.8} 
& \bl{+3.0} & \bl{+3.2} & \bl{+3.2} & \bl{+0.2} 
& \bl{+5.1} & \bl{+3.5}
\\ \hline

Qwen2-VL-7B      & 62.0 & 70.8 & 41.4  & 68.7 & 64.6 & 58.4 & 80.9 & 90.0  & 63.2 & 66.6 \\ 
Qwen2-VL-7B w/ BDR     & \textbf{64.3} & \textbf{73.1} & \textbf{43.5} & \textbf{69.9} & \textbf{66.7} & \textbf{60.5} & \textbf{83.4} & \textbf{91.0} & \textbf{65.3} & \textbf{68.6} \\
\color{blue}{$\Delta$}                  
& \bl{+2.3} & \bl{+2.3} & \bl{+2.1} & \bl{+1.2} 
& \bl{+2.1} & \bl{+2.1} & \bl{+2.5} & \bl{+1.0} 
& \bl{+2.1} & \bl{+2.0}
\\
\bottomrule
\end{tabular}
\label{tab:m2kr-results}
\end{table*}

\begin{table*}[htb]
\centering
\caption{
\textbf{Answer generation performance comparison on InfoSeek and EVQA.} 
We report VQA Accuracy, Exact Match (EM), BLEU-1, and \bl{BERT Matching (BEM)}. 
The Oracle Retriever retrieves all ground-truth documents. 
Our BDR retriever consistently outperforms PreFLMR \cite{preflmr} and ReT \cite{recurrence}, approaching the Oracle upper bound.
}
\setlength{\tabcolsep}{6pt}
\renewcommand{\arraystretch}{0.85}
\fontsize{8}{8}\selectfont
\renewcommand{\arraystretch}{1.1}
\begin{tabular}{llc|ccc|cccc}
\toprule
\multicolumn{1}{l}{}   
& \multicolumn{1}{l}{} 
& \multicolumn{1}{l}{}       
& \multicolumn{3}{c}{\textbf{InfoSeek}}                  
& \multicolumn{4}{c}{\textbf{EVQA}} \\ 
\hline

\textbf{Generator (Frozen)} 
& \textbf{Retriever} 
& \textbf{R@5} 
& \textbf{VQA\_Acc} & \textbf{EM} & \textbf{BLEU\_1} 
& \textbf{R@5} 
& \textbf{VQA\_Acc} & \textbf{BLEU\_1} & \bl{\textbf{BEM}} \\

LLaVA-1.6-13B          
& \ding{55}                   
& - & 5.4 & 5.3 & 11.9    
& - & 2.7 & 8.9 & \bl{69.8} \\

LLaVA-1.6-13B          
& PreFLMR              
& 39.2 & 12.9 & 12.4 & 21.2 
& 62.0 & 8.7 & 26.2 & \bl{74.3} \\

LLaVA-1.6-13B          
& ReT                  
& 52.0 & 17.3 & 17.2 & 28.9 
& 48.6 & 6.5 & 19.2 & \bl{73.2} \\

LLaVA-1.6-13B          
& Qwen2-VL-7B (Ours)          
& \textbf{66.7} & \textbf{20.8} & \textbf{20.9} & \textbf{34.0} 
& \textbf{64.3} & \textbf{9.1} & \textbf{26.9} & \bl{\textbf{77.2}} \\

LLaVA-1.6-13B          
& Oracle Retriever     
& - & 37.5 & 39.5 & 56.4 
& - & 16.1 & 46.1 & \bl{86.7} \\ 

\midrule

Qwen2.5-VL-7B          
& \ding{55}                   
& - & 14.4 & 14.5 & 25.2 
& - & 4.6 & 14.3 & \bl{65.2} \\

Qwen2.5-VL-7B          
& PreFLMR              
& 39.2 & 21.5 & 16.1 & 24.1 
& 62.0 & 11.5 & 34.7 & \bl{68.3} \\

Qwen2.5-VL-7B          
& ReT                  
& 52.0 & 25.9 & 21.5 & 32.2 
& 48.6 & 10.8 & 28.1 & \bl{67.9} \\

Qwen2.5-VL-7B          
& Qwen2-VL-7B (Ours)             
& \textbf{66.7} & \textbf{32.1} & \textbf{27.5} & \textbf{41.3} 
& \textbf{64.3} & \textbf{14.4} & \textbf{37.1} & \bl{\textbf{71.2}} \\

Qwen2.5-VL-7B          
& Oracle Retriever     
& - & 46.2 & 41.3 & 61.9 
& - & 23.3 & 57.8 & \bl{89.1} \\ 

\bottomrule
\end{tabular}
\label{tab:answer-comparison}
\end{table*}

\section{Experiments}
In the experimental section, we investigate five key research questions: 
\textbf{RQ1:} How does Bayesian Data Reweighting compare with existing reweighting methods?
\textbf{RQ2:} How does the reweighting improve downstream VQA performance?
\textbf{RQ3:} How does Bayesian Data Reweighting adjust sample importance?
\textbf{RQ4:} How should the Bayesian prior hyperparameters be determined?
\textbf{RQ5:} Does it introduce additional computational overhead?

\subsection{Comparison with Existing Baselines}
We consider the following reweighting-based methods as baselines.
\textbf{Random Negative Reweighting} assigns stochastic weights on negative samples randomly;
\textbf{Similarity-based Reweighting}~\cite{sampling} emphasizes high-similarity negatives using
\(w_{ik}^{-}\propto \exp(\gamma s_{ik}^{-})\);
\textbf{Margin-based Reweighting}~\cite{facenet} focuses on a certain margin gap by enforcing
\(\mathcal{L}_{\mathrm{margin},i}=[\alpha+s_{ik}^{-}-s_i^{+}]_{+}\);
\textbf{Decoupled Positive Reweighting}~\cite{decoupled} removes the positive pair from the denominator and upweights hard positives, 
\(\mathcal{L}_{\mathrm{DCLW},i}=-w_i s_i^{+}/\tau+\log\sum_{k}\exp(s_{ik}^{-}/\tau)\);
\textbf{Debiased Negative Reweighting}~\cite{chuang2020debiased} corrects false-negative bias by estimating the true-negative contribution;
and \textbf{Hardness Negative Reweighting}~\cite{hard-negative} assigns hardness-aware weights
\(q_{ik}^{-}\propto \exp(\beta s_{ik}^{-})\) and combines them with the false-negative debiasing.

\paragraph{Experimental Setup.}
To compare with existing reweighting methods, we conduct experiments on OKVQA \cite{okvqa} and EVQA \cite{commonsensevqa}. For each dataset, we sample 10K training examples to fine-tune a multimodal LLM-based embedding model and evaluate retrieval performance on the test set. Note the original evaluation corpus contains millions of documents, so we filter the test corpus to retain only ground-truth-contained documents for faster retrieval evaluation.
We report Recall@K (R@K), which measures whether the target document appears in the top-$K$ retrieved results, and Pseudo Recall@K (PR@K), which measures whether any top-$K$ document contains the correct answer, following prior work \cite{rag-vqa, preflmr}.

We use Qwen2-VL-2B~\cite{qwen2-VL} as the backbone and apply LoRA-based fine-tuning with different reweighting strategies. Following VLM2Vec~\cite{vlm2vec}, the model uses the last token as the input embedding for contrastive learning. The main hyperparameters are: LoRA rank $r=4$, LoRA alpha $=64$, batch size $=64$, learning rate $=1\times10^{-4}$, dropout $=0.1$, and temperature $\tau=0.02$. All models are trained on 4 NVIDIA A100 GPUs under the same optimization setup. Results are reported in Table~\ref{tab:reweighting_r1_r5_pr1_pr5}.

\paragraph{Reweighting Results Analysis.}
As shown in Table~\ref{tab:reweighting_r1_r5_pr1_pr5}, Bayesian reweighting achieves the best performance on both OKVQA and EVQA across all metrics. Among the baselines, hardness negative reweighting and debiased negative reweighting perform competitively, suggesting that false-negative mitigation is important for KB-VQA retrieval. In contrast, heuristic or deterministic methods, such as random, similarity-based, margin-based, provide limited gains because they rely on fixed rules or sensitive hyperparameters, therefore limiting their generalization ability. BDR instead models sample importance as latent variables and adaptively infers posterior weights, yielding more consistent improvements. The Gamma-prior variant further outperforms the Bernoulli-prior variant, indicating that continuous uncertainty-aware reweighting is more effective than binary selection for diverse negative samples.

\paragraph{Broader M2KR Benchmarks.}
We further evaluate BDR on seven M2KR datasets~\cite{preflmr} using Qwen2-VL-2B as the backbone, with 1K training samples per task trained jointly. As shown in Table~\ref{tab:multiple_benchmarks}, BDR achieves the best average retrieval performance, with the Gamma-prior variant performing best overall and the Bernoulli-prior variant achieving strong results on InfoSeek \cite{infoseek}. These results demonstrate that BDR generalizes well across diverse knowledge-intensive VQA benchmarks and consistently improves over existing reweighting baselines.

\paragraph{Broader Retriever Backbones.}
To evaluate the general applicability of BDR, we apply the Gamma-prior variant to three LLM backbones, including Qwen2-VL-2B, Phi-3.5-V-3.8B, and Qwen2-VL-7B, using the complete M2KR training corpus and evaluating on the full test set. As shown in Table~\ref{tab:m2kr-results}, BDR achieves strong performance across different model scales and surpasses previous state-of-the-art retrievers on this benchmark, including Pre-FLMR~\cite{preflmr} and ReT~\cite{recurrence}. Notably, Qwen2-VL-7B w/ BDR obtains the best overall average performance, demonstrating that BDR is broadly effective across diverse backbone architectures for knowledge-intensive VQA retrieval.

\vspace{-10pt}
\subsection{Impact on Downstream VQA Task}
\textbf{Answer generation results.}
To evaluate downstream VQA performance, we integrate the trained retriever with two frozen multimodal LLMs, LLaVA-1.6-13B \cite{llava-next} and Qwen2.5-VL-7B \cite{qwen2.5-VL}, to build a complete retrieval-augmented VQA pipeline. We evaluate the generated answers on InfoSeek and EVQA using VQA Score \cite{rag-vqa} and related generation metrics. As shown in Table~\ref{tab:answer-comparison}, the BDR-based retriever consistently achieves the best answer generation performance across both datasets and generators. Compared with existing strong retrievers, BDR improves VQA accuracy, exact match, BLEU-1, and BEM in most cases, indicating that better retrieval quality effectively translates into stronger downstream reasoning. Moreover, its performance is closer to the Oracle Retriever, suggesting that BDR retrieves more relevant evidence and better narrows the gap between practical retrieval and oracle-level generation.

\begin{figure}[htbp]
    \includegraphics[width=0.475\textwidth]{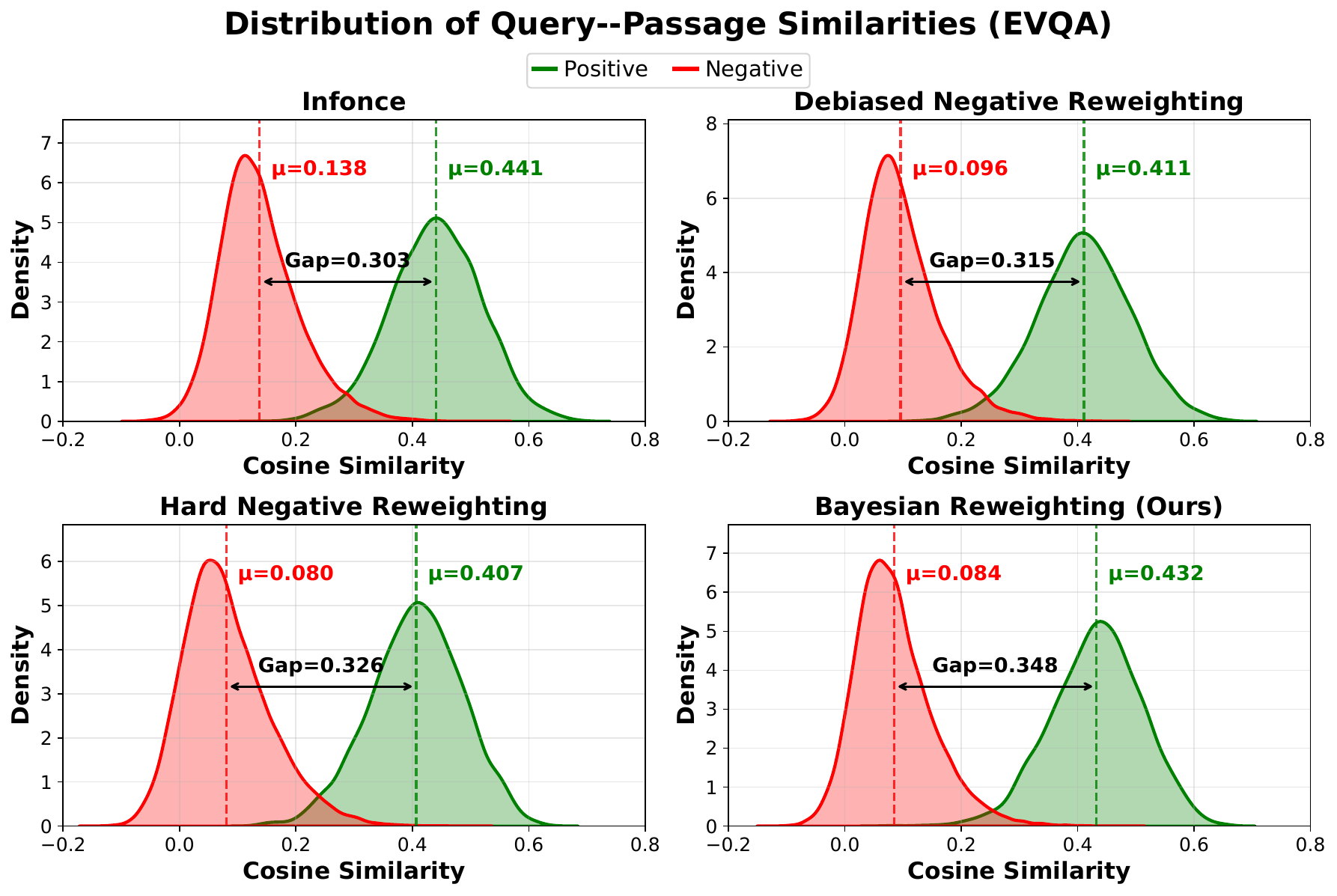}
    \caption{
    \textbf{Distribution of query--passage cosine similarities on EVQA.}
    The dashed lines denote the mean of similarities, and the annotated gap measures the positive--negative separation. 
    }
    
\label{fig:evqa_similarity_distribution}
\end{figure}

\begin{figure}[htbp]
    \includegraphics[width=0.475\textwidth]{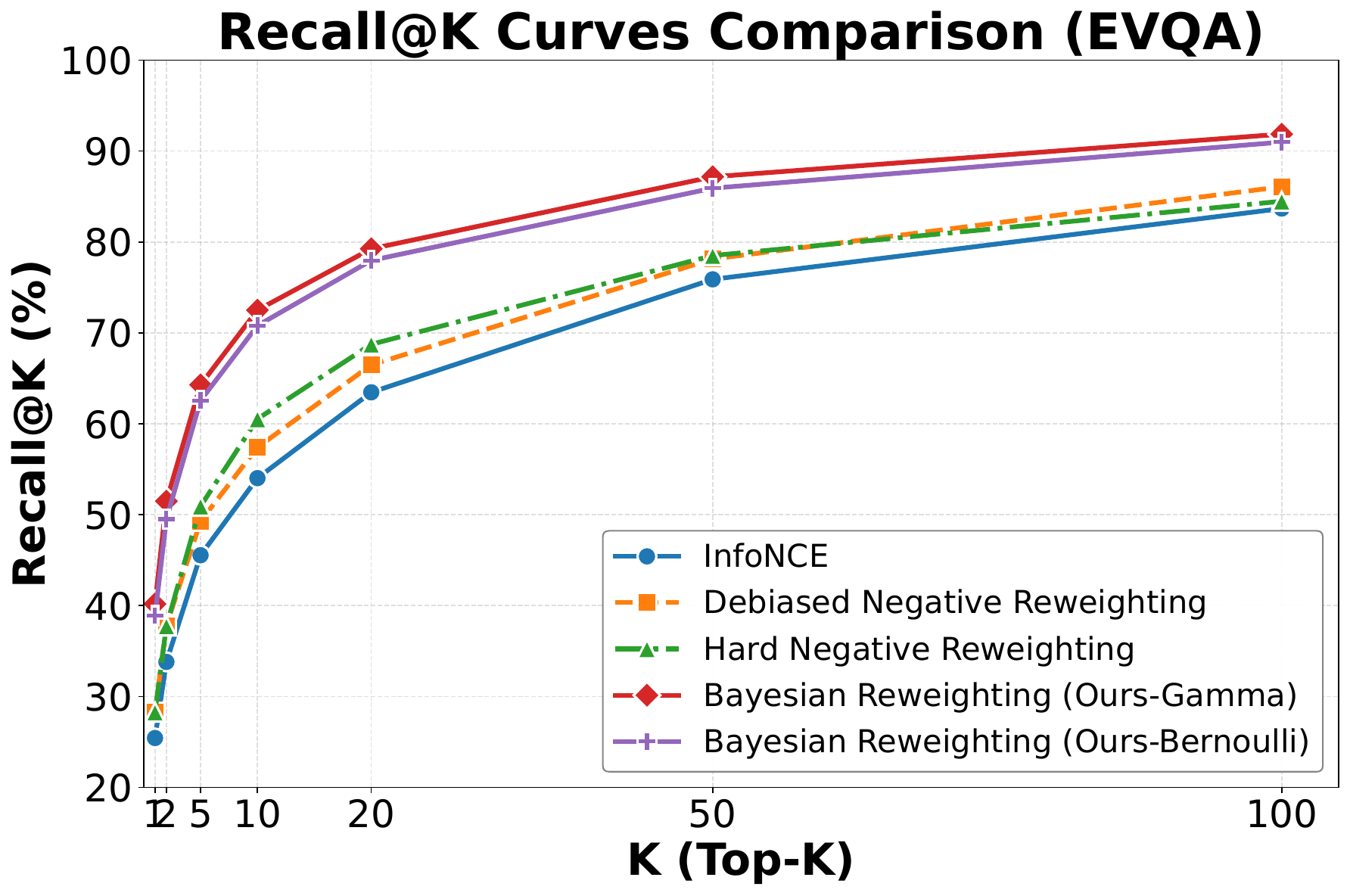}
    \caption{\textbf{Recall@K comparison on EVQA.} We evaluate retrieval performance by varying $K$ and reporting whether the ground-truth document appears in the top-$K$ results. }
    \vspace{-20pt}
    \label{fig:recall_curve}
\end{figure}

\subsection{How does Bayesian Reweighting work?}
\textbf{Decision Boundary Visualization.} 
We analyze the cosine similarity distributions of positive and negative query--passage pairs to examine the effect of Bayesian Reweighting on the decision boundary. As shown in Figure~\ref{fig:evqa_similarity_distribution}, our method achieves the largest positive--negative similarity gap, improving it from \(0.303\) with InfoNCE to \(0.348\). This gap is also larger than those of debiased negative reweighting and hardness negative reweighting, indicating that Bayesian Reweighting learns more discriminative embeddings by better separating relevant and irrelevant passages.

\textbf{Recall@K analysis.}
We analyze how the performance of retrievers trained with different methods changes by varying the number of top-\(K\) retrieved documents. The results are shown in Figure~\ref{fig:recall_curve}. Compared with retrievers trained using other reweighting methods, the two retrievers trained with BDR achieve consistently higher Recall@\(K\) across different top-\(K\) settings, clearly outperforming existing baselines. The gains are especially pronounced at smaller \(K\), indicating that BDR improves the ranking of relevant documents near the top of the retrieved list. This experiment demonstrates that BDR yields more robust retrieval quality across different retrieval budgets.

\begin{figure}[t]
    \centering
    \includegraphics[width=\linewidth]{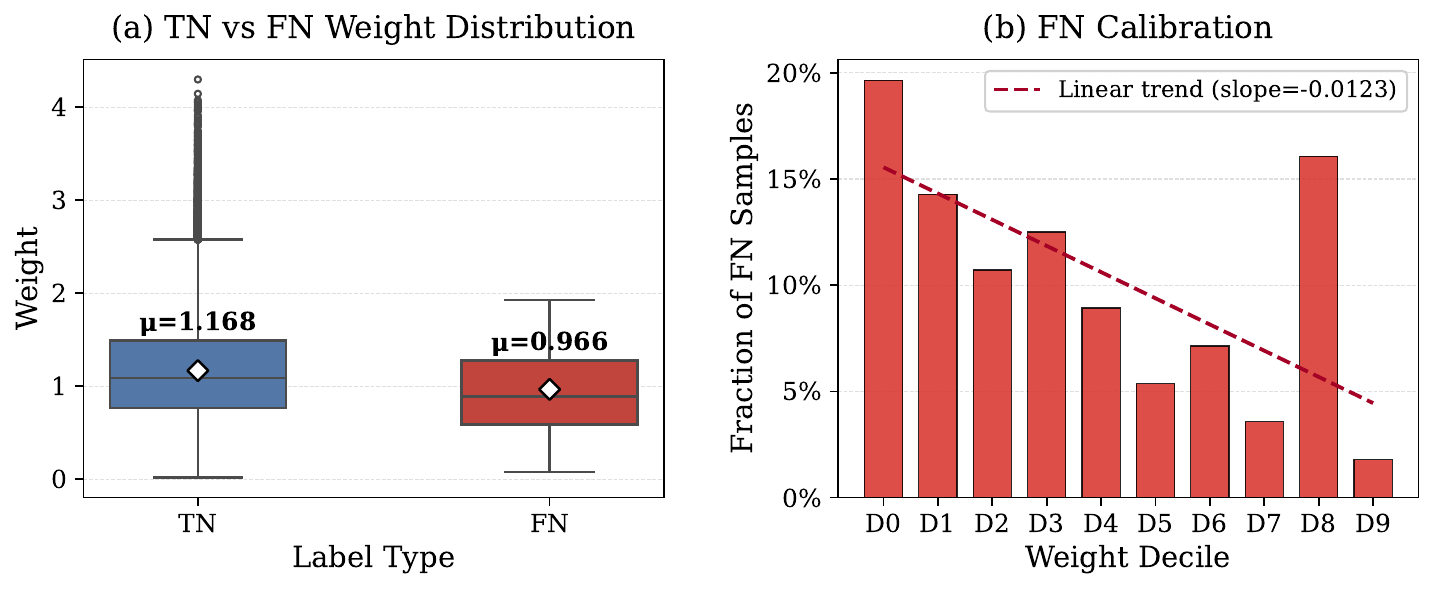}
    \caption{\textbf{Weight calibration analysis.} Left: FN samples receive lower Bayesian weights than TN samples on average. Right: the FN fraction generally decreases across increasing weight deciles, suggesting that lower-weight samples are more likely to be false negatives.}
    \label{fig:weight_calibration}
    \vspace{-15pt}
\end{figure}

\textbf{Weight Calibration Analysis.}
To evaluate whether the learned Bayesian weights align with sample quality, we conduct a calibration analysis on EVQA. We first use an external LLM embedding model to measure passage-level semantic similarity, followed by manual verification to identify false negatives (FNs). Among 1,000 sampled query--passage pairs, 54 are identified as FNs. We then record the weights inferred by BDR, divide negative samples into ten equal-frequency weight deciles, and compute the FN proportion in each bin. As shown in the right panel of Figure~\ref{fig:weight_calibration}, the FN proportion decreases as the weight decile increases, indicating that lower-weight bins contain more FNs while higher-weight bins are dominated by true negatives (TNs). We also visualize the weight distributions of TNs and FNs, where FN samples receive lower weights on average than TN samples, suggesting that BDR effectively suppresses potentially mislabeled negatives. These results demonstrate that the Bayesian weighting mechanism is well calibrated with respect to sample quality.

\vspace{-5pt}
\subsection{Sensitivity to Bayesian Priors}
We study the sensitivity of BDR to different prior hyperparameters on OKVQA. For the Gamma prior, we fix $a_w=a_u=1$ and vary $b_w$ and $b_u$ to examine the effect of prior strength. As shown in Tables~\ref{tab:sensitivity_bw_bu_okvqa} , BDR remains stable across most settings, while an overly large $b_u$ degrades performance, indicating that too strong a prior may limit adaptive weight inference. For the Bernoulli prior, we vary the retaining probability $p$ and observe only mild performance changes, Table~\ref{tab:sensitivity_p_n_okvqa} suggesting that binary gating is also reasonably robust. We further test the number of posterior sampling iterations $N$ under the Gamma-prior setting. Performance remains stable from $N=1$ to $8$, showing that BDR can achieve reliable posterior estimates with only a few sampling steps.

\begin{table}[t]
\centering
\caption{Sensitivity analysis of $b_w$ and $b_u$ on OKVQA.}
\label{tab:sensitivity_bw_bu_okvqa}
\setlength{\tabcolsep}{3pt}
\fontsize{7}{7}\selectfont
\renewcommand{\arraystretch}{1}
\begin{tabular}{c|cccc|c|cccc}
\toprule
\multicolumn{5}{c|}{\textbf{$b_w$}} & \multicolumn{5}{c}{\textbf{$b_u$}} \\
\midrule
\textbf{$b_w$} & R@1 & R@5 & PR@1 & PR@5 
&\textbf{$b_u$}  & R@1 & R@5 & PR@1 & PR@5 \\
\midrule
0.1 & 30.0 & 54.8 & 64.3 & 65.7 
& 0.1  & 30.1 & 54.8 & 66.0 & 66.1 \\
0.5 & 27.9 & 52.5 & 63.9 & 63.5 
& 0.5  & 29.9 & 54.5 & 65.2 & 64.7 \\
1.0 & 28.6 & 53.2 & 63.6 & 63.3 
& 1.0  & 29.1 & 54.4 & 64.7 & 65.3 \\
2.0 & 27.6 & 53.1 & 63.9 & 63.2 
& 5.0  & 29.9 & 53.7 & 66.0 & 63.9 \\
5.0 & 28.7 & 53.5 & 64.4 & 64.4 
& 20.0 & 26.1 & 50.4 & 62.8 & 60.7 \\
\bottomrule
\end{tabular}
\end{table}

\begin{table}[t]
\centering
\caption{Sensitivity analysis of $p$ and $N$ on OKVQA.}
\label{tab:sensitivity_p_n_okvqa}
\setlength{\tabcolsep}{3pt}
\fontsize{7}{7}\selectfont
\renewcommand{\arraystretch}{1}
\begin{tabular}{c|cccc|c|cccc}
\toprule
\multicolumn{5}{c|}{\textbf{$p$}} & \multicolumn{5}{c}{\textbf{$N$}} \\
\midrule
$p$ & R@1 & R@5 & PR@1 & PR@5 
&$N$  & R@1 & R@5 & PR@1 & PR@5 \\
\midrule
0.50 & 28.68 & 53.17 & 63.69 & 64.21 
& 1 & 27.90 & 53.11 & 65.42 & 63.89 \\
0.70 & 28.70 & 52.81 & 64.59 & 63.54 
& 2 & 28.97 & 54.02 & 64.17 & 64.76 \\
0.85 & 28.76 & 53.84 & 64.39 & 64.96 
& 4 & 28.76 & 53.73 & 63.30 & 63.81 \\
0.95 & 29.01 & 54.44 & 64.92 & 65.72 
& 8 & 29.07 & 54.06 & 64.17 & 65.50 \\
\bottomrule
\end{tabular}
\end{table}

\vspace{-5pt}
\subsection{Training Efficiency Comparison}
Bayesian Reweighting is computationally efficient because its EM updates only require closed-form, element-wise sampling with a lightweight cost of $O(MBK)$. Since the dominant cost comes from the multimodal encoder forward and backward updates, the additional reweighting overhead is negligible in practice. As shown in Table~\ref{tab:training-efficiency}, BDR achieves nearly identical GPU memory usage and training time compared with standard InfoNCE across different backbones and batch sizes. 

\begin{table}[htbp]
\caption{\bl{\textbf{Model training efficiency comparison.}}}
\label{tab:training-efficiency}
\setlength{\tabcolsep}{7pt}
\fontsize{7}{7}\selectfont
\renewcommand{\arraystretch}{0.8}
\centering
\begin{tabular}{llccc}
\toprule
\multicolumn{2}{c}{} & \multicolumn{3}{c}{\textbf{\bl{Batch Size = 32}}} \\
\cmidrule(lr){3-5}
\bl{Models} & \bl{Method} & \bl{Steps} & \bl{GPU (GB)} & \bl{Time (h)} \\
\midrule
\bl{Phi-3.5-V}    & \bl{InfoNCE}    & \bl{521} & \bl{19.9} & \bl{3.17} \\
\bl{Phi-3.5-V}    & \bl{BDR (Ours)} & \bl{521} & \bl{20.0} & \bl{3.22} \\
\bl{Qwen2-VL-7B}  & \bl{InfoNCE}    & \bl{521} & \bl{32.8} & \bl{2.01} \\
\bl{Qwen2-VL-7B}  & \bl{BDR (Ours)} & \bl{521} & \bl{32.9} & \bl{2.05} \\
\midrule
\multicolumn{2}{c}{} & \multicolumn{3}{c}{\textbf{\bl{Batch Size = 128}}} \\
\cmidrule(lr){3-5}
\bl{Models} & \bl{Method} & \bl{Steps} & \bl{GPU (GB)} & \bl{Time (h)} \\
\midrule
\bl{Phi-3.5-V}    & \bl{InfoNCE}    & \bl{131} & \bl{19.8} & \bl{3.16} \\
\bl{Phi-3.5-V}    & \bl{BDR (Ours)} & \bl{131} & \bl{20.3} & \bl{3.18} \\
\bl{Qwen2-VL-7B}  & \bl{InfoNCE}    & \bl{131} & \bl{32.8} & \bl{2.00} \\
\bl{Qwen2-VL-7B}  & \bl{BDR (Ours)} & \bl{131} & \bl{32.9} & \bl{2.02} \\
\bottomrule
\end{tabular}
\vspace{-10pt}
\end{table}

\vspace{-5pt}
\section{Conclusion}

We propose Bayesian Data Reweighting (BDR), a principled and efficient framework for uncertainty-aware contrastive learning in knowledge-intensive VQA retrieval. BDR adaptively reweights training pairs, consistently improving retrieval performance across M2KR datasets and multiple LLM backbones, which further benefits downstream answer generation. Analysis shows that BDR learns more discriminative embeddings, suppresses harmful false negatives, and adds negligible computational overhead. Future work will extend BDR beyond retrieval-based VQA to broader multimodal RAG and reasoning tasks.

\section*{Limitations}
This work focuses on knowledge-based VQA retrieval, although BDR is evaluated across multiple KB-VQA benchmarks and retriever backbones, it has several limitations. First, the evaluation is limited to the datasets, models, and metrics considered in this paper, and BDR is not directly designed for multi-hop retrieval. Second, from a strict Bayesian perspective, the posterior is formulated using a pseudo-likelihood for tractable derivation. Therefore, it is not a fully normalized posterior distribution, but serve as uncertainty-aware importance weights. Future work will validate BDR under full open-corpus retrieval and explore joint retriever-generator optimization for end-to-end multimodal RAG systems.

\section*{Ethical Considerations}

This work aims to improve multimodal retrieval for knowledge-based VQA and does not introduce new datasets or collect personal information. However, retrieval-augmented systems may still retrieve biased, outdated, or misleading evidence, which can lead to incorrect downstream answers. Therefore, BDR should not be viewed as a guarantee of factuality or safety, especially in high-stakes applications where human verification remains necessary.
\bibliography{custom}

\appendix



\end{document}